\documentclass[conference]{IEEEtran}
\usepackage{cite}
\usepackage{amsmath,amssymb,amsfonts}
\usepackage{algorithmic}
\usepackage{graphicx}
\usepackage{textcomp}
\usepackage{xcolor}
\usepackage{indentfirst}
\usepackage{float}
\usepackage{url}

\def\BibTeX{{\rm B\kern-.05em{\sc i\kern-.025em b}\kern-.08em
    T\kern-.1667em\lower.7ex\hbox{E}\kern-.125emX}}
\begin{document}

\title{Bayesian Monocular Depth Refinement via\\Neural Radiance Fields}

\author{\IEEEauthorblockN{Arun Muthukkumar}
\IEEEauthorblockA{\textit{Department of Computer Science} \\
\textit{Illinois Mathematics and Science Academy}\\
Aurora, United States \\
amuthukkumar@imsa.edu}
}

\maketitle
\begingroup
\renewcommand\thefootnote{}\footnotetext{\copyright\ 2026 IEEE. Accepted for publication in the 2025 8th International Conference on Algorithms, Computing and Artificial Intelligence (ACAI). Personal use is permitted. All other uses require permission from IEEE.}
\endgroup

\begin{abstract}

Monocular depth estimation has applications in many fields, such as autonomous navigation and extended reality, making it an essential computer vision task. However, current methods often produce smooth depth maps that lack the fine geometric detail needed for accurate scene understanding. We propose MDENeRF, an iterative framework that refines monocular depth estimates using depth information from Neural Radiance Fields (NeRFs). MDENeRF consists of three components: (1) an initial monocular estimate for global structure, (2) a NeRF trained on perturbed viewpoints, with per-pixel uncertainty, and (3) Bayesian fusion of the noisy monocular and NeRF depths. We derive NeRF uncertainty from the volume rendering process to iteratively inject high-frequency fine details. Meanwhile, our monocular prior maintains global structure. We demonstrate improvements on key metrics and experiments using indoor scenes from the SUN RGB-D dataset.

\end{abstract}

\setlength{\parindent}{20pt}

\section{Introduction}
\label{sec:intro}

Monocular depth estimation (MDE) is an ill-posed problem. Learning-based approaches are capable of recovering global structures, but they often struggle on thin objects and sharp depth discontinuities~\cite{ming2021deep}. These are crucial limitations to overcome, as poor scene understanding can be a bottleneck for downstream applications such as robotics and augmented reality.

Neural rendering provides complementary geometric cues. Neural Radiance Fields (NeRFs) excel at synthesizing locally consistent novel viewpoints and learning the scene geometry implicitly through volumetric density~\cite{mildenhall2021}. A confident NeRF inference can aid MDE, which naturally leads to a \emph{probabilistic refinement approach} that leverages nearby viewpoints, even with only a single available image.

\begin{figure*}
  \centering
  \includegraphics[width=.8\linewidth]{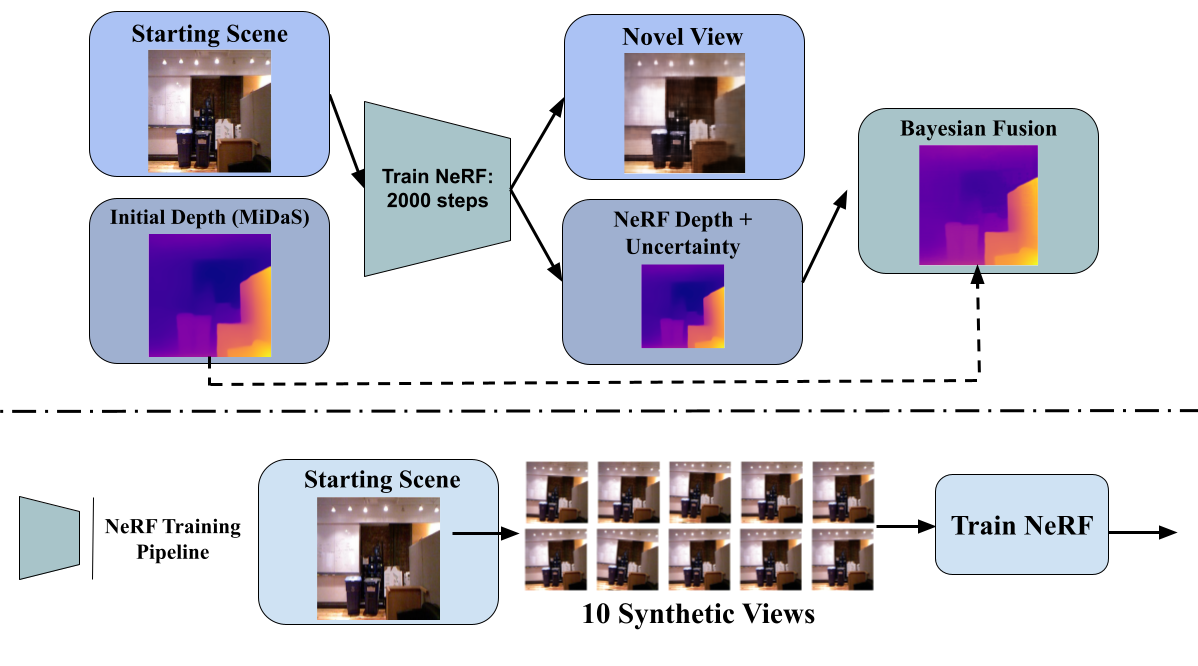}
  \caption{One MDENeRF refinement iteration. A NeRF trains on nearby synthetic views and renders depth and uncertainty. NeRF and monocular depth estimates are fused via Bayesian inference. The process repeats for 2–3 iterations, which progressively enhances detail.
}
  \label{fig:pipe}
\end{figure*}

We propose \textbf{MDENeRF}, an iterative depth refinement framework that improves our initial monocular depth estimate (See Figure~\ref{fig:pipe}). First, we synthetically create nearby viewpoints using perturbed images. These simulate a multi-view environment that enhances NeRF training. The depth learned from the NeRF can be reprojected to the original pose, and we statistically formulate the \emph{per-pixel uncertainty} of the resulting NeRF depth map. Both our monocular estimate and our trained NeRFs are noisy, so we use \emph{Bayesian inference} to fuse the two estimates based on their uncertainty. We find that this preserves global structure during refinement while iteratively refining high-frequency details. Quantitative and qualitative evaluations demonstrate robust depth refinement. Code is available at \url{https://github.com/ArunMut/MDENeRF}.

\section{Related Work}
\label{sec:related}

\subsection{Monocular Depth Estimation}
Over the last decade, MDE has undergone significant advancements. Early supervised methods relied on multi-scale CNNs to predict depth from a single image. Eigen et al.\ pioneered coarse-to-fine architectures to learn global and local depth in indoor scenes~\cite{eigen2014}. Liu et al.\ enhanced spatial coherence by integrating continuous CRFs within CNNs~\cite{liu2015}. Laina et al.\ further enhanced accuracy via deep residual networks with up-projection blocks~\cite{laina2016}. Despite improvements, MDE depth maps remained overly smooth and frequently suppressed fine structures.

To mitigate reliance on ground-truth depth, methods shifted toward self-supervised and unsupervised learning. Zhou et al.\ jointly learned depth and ego-motion from video with view synthesis for supervision~\cite{zhou2017}. MonoDepth introduced left-right consistency for stereo-based training~\cite{godard2017}. MonoDepth2 later extended to monocular video with auto-masking and multi-scale losses~\cite{godard2019}. Kendall et al.\ incorporated geometric cues into end-to-end stereo disparity networks~\cite{kendall2017}.

Recent methods have scaled to larger datasets and architectures. MiDaS was trained on multiple datasets to ensure strong cross-dataset generalization and global consistency~\cite{ranftl2022}. To capture high-frequency details, DORN formulated depth estimation as ordinal regression, and AdaBins utilized adaptive binning with transformer backbones to preserve thin structures~\cite{bhat2021}. Despite these advances, state-of-the-art MDE still struggles with thin objects, occlusion boundaries, and fine geometric detail, particularly in challenging indoor scenes~\cite{obukhov2025fourth}.

\subsection{Neural Rendering and Depth Estimation}

Neural Radiance Fields (NeRFs) represent scenes as continuous volumetric functions, which allow for photorealistic novel view synthesis through differentiable rendering~\cite{mildenhall2021}. Depth emerges from NeRF as the expected ray termination distance under the learned volumetric density. This provides geometrically meaningful estimates that are often sharper at boundaries than those from MDE~\cite{uy2023scade}. Subsequent work improved NeRF efficiency and quality through multi-resolution hash encodings, consistency constraints, and geometric regularization~\cite{muller2022}.

Several works have explored combining NeRFs with depth estimation. Parallel works inject depth priors during training or use NeRF-generated views to augment datasets~\cite{feldmann2024nerfmentation}. In contrast, our approach applies NeRFs at \emph{test time} to \emph{refine} a single-view depth estimate. Crucially, MDENeRF probabilistically formulates NeRF volume rendering to derive per-pixel depth uncertainty in closed form, which guides principled fusion with monocular depth.

\subsection{Depth Map Fusion and Refinement}

Depth refinement has been addressed through post-processing filters and learned propagation mechanisms, such as convolutional spatial propagation networks (CSPN)~\cite{cheng2019}, as well as losses that align depth and image gradients to sharpen edges~\cite{ramamonjisoa2020}. More recent work explores fusing multiple depth sources to recover fine detail while preserving global coherence~\cite{dai2023}.

MDENeRF falls into this latter category but differs fundamentally in formulation. Instead of heuristic or gradient-based blending, we model monocular and NeRF depth as noisy observations of the same underlying scene geometry. This allows us to fuse them via Bayesian inference. Our uncertainty enables selective refinement driven by NeRF confidence, preserves global structure, and avoids hand-tuned fusion parameters.

\section{MDENeRF: Methodology}

MDENeRF operates under the assumption that we are limited to a single RGB image of a static scene and a monocular depth estimator. Ground-truth depth is a latent variable that we observe through two noisy sources: MDE and NeRFs. MDENeRF uses Bayesian fusion of monocular and NeRF depth, effectively injecting locally consistent geometry from nearby viewpoints while preserving the monocular prior. 

\subsection{Synthetic Data Generation}
A setting with $N$ views can be simulated from one RGB image with small, controlled camera perturbations (a few degrees/centimeters) about the optical center. Each of these perturbations defines a pose change. We warp the original image and depth map accordingly, producing our pseudo multi-view dataset
\begin{equation}
\bigl(I_{\text{orig}}, D_{\text{orig}}\bigr)\;\;\;\text{and}\;\;\;\Bigl\{\bigl(I_{\text{syn}}^i, D_{\text{syn}}^i\bigr)\Bigr\}_{i=1}^{N}.
\end{equation}
Later, we reproject $D_{\text{syn}}^i$ to the original pose. We limit our synthetic dataset to mild perturbations to mitigate unrealistic artifacts. This dataset is used for training the NeRFs.

\subsection{Initial Depth Estimation}
We define $D_o$ as our initial depth map, provided by a monocular depth estimator. As explained in Sec.~\ref{sec:related}, $D_o$ is a spatially smooth coarse estimate, motivating our use of multi-view geometric cues.

\subsection{NeRF Depth and Uncertainty}
NeRFs model radiance and density along each ray $\mathbf{r}(t)=\mathbf{o}+t\mathbf{d}$. Sampled points $\{\mathbf{r}(t_i)\}_{i=1}^{M}$ with predicted densities $\sigma_i$ are used to define the opacity and accumulated transmittance in standard volume rendering:
\begin{equation}
\alpha_i = 1 - \exp(-\sigma_i \Delta_i), \qquad
T_i = \prod_{j<i} (1-\alpha_j).
\end{equation}
Conditional on ray termination---which holds for the vast majority of rays in bounded indoor scenes---ray termination weights form a discrete probability distribution:
\begin{equation}
w_i = T_i \alpha_i, \qquad
p_i = \frac{w_i}{\sum_k w_k}.
\end{equation}

The rendered depth is the expected termination:
\begin{equation}
\mu_r(\mathbf{r}) = \sum_{i=1}^{M} p_i\, t_i.
\label{eq:nerf_mean}
\end{equation}
The second moment and variance are then
\begin{equation}
m_2(\mathbf{r}) = \sum_{i=1}^{M} p_i\, t_i^2, \qquad
\sigma_r^2(\mathbf{r}) = m_2(\mathbf{r}) - \mu_r(\mathbf{r})^2.
\label{eq:nerf_var}
\end{equation}
Both processes are principled and computationally inexpensive, as we achieve per-ray uncertainty with the same rendering weights.

\subsection{Novel View Synthesis and Depth Reprojection}
After training, our NeRF module renders a small set of novel views $j\in\{1,\dots,K\}$. As per Eqs.~\eqref{eq:nerf_mean}--\eqref{eq:nerf_var}, we obtain per-pixel mean NeRF depth $\mu_r^{(j)}$ and variance $(\sigma_r^2)^{(j)}$ for each rendered view. Next, we project each depth map into the original camera frame. For a pixel $x$ in this original view, let $\mathcal{V}(x)$ denote the set of rendered views that project validly to $x$, yielding estimates
\begin{equation}
\Bigl\{ \mu_r^{(j)}(x),\, (\sigma_r^2)^{(j)}(x)\Bigr\}_{j\in\mathcal{V}(x)}.
\end{equation}

Rather than heuristic aggregation, we use precision weighting to fuse the $K$ projected NeRF predictions as Gaussians:
\begin{equation}
\begin{aligned}
\tau_r(x) &= \sum_{j\in\mathcal{V}(x)}
\frac{1}{(\sigma_r^2)^{(j)}(x)}, \\[4pt]
\mu_r^{\text{agg}}(x) &= \frac{1}{\tau_r(x)}
\sum_{j\in\mathcal{V}(x)}
\frac{\mu_r^{(j)}(x)}{(\sigma_r^2)^{(j)}(x)}.
\end{aligned}
\end{equation}
\begin{equation}
(\sigma_r^2)^{\text{agg}}(x) = \frac{1}{\tau_r(x)}.
\label{eq:mv_nerf_fuse}
\end{equation}
If $\mathcal{V}(x)=\emptyset$ (no valid reprojection), we set $\tau_r(x)=0$ and treat the NeRF as missing at $x$.

\subsection{Bayesian Depth Fusion}
We fuse monocular depth $D_o$ with the consolidated NeRF depth $\mu_r^{\text{agg}}$ by treating each as a noisy observation of the unknown true depth $D(x)$:
\begin{equation}
D_o(x) = D(x) + \varepsilon_o(x),\qquad
\mu_r^{\text{agg}}(x) = D(x) + \varepsilon_r(x),
\end{equation}
with $\varepsilon_o(x)\sim\mathcal{N}(0,\sigma_o^2)$ and $\varepsilon_r(x)\sim\mathcal{N}\!\bigl(0,(\sigma_r^2)^{\text{agg}}(x)\bigr)$. Here $(\sigma_r^2)^{\text{agg}}(x)$ is computed from the NeRF (Eq.~\eqref{eq:mv_nerf_fuse}), and $\sigma_o^2$ is estimated from the scene via empirical Bayes (below). This allows for fusion without hyperparameters.

Because monocular depth is scale-ambiguous, we align NeRF depth to the monocular scale using a weighted affine mapping
\begin{equation}
\tilde{D}_r(x) = a\,\mu_r^{\text{agg}}(x) + b,
\label{eq:affine}
\end{equation}
where $(a,b)$ minimize weighted squared error over pixels with valid NeRF aggregation:
\begin{equation}
(a^\star,b^\star)=\arg\min_{a,b}\;\sum_{x\in\mathcal{X}}\frac{\bigl(a\,\mu_r^{\text{agg}}(x)+b-D_o(x)\bigr)^2}{(\sigma_r^2)^{\text{agg}}(x)},
\label{eq:wls}
\end{equation}
and $\mathcal{X}=\{x:\tau_r(x)>0\}$. Let $u(x)=\mu_r^{\text{agg}}(x)$, $v(x)=D_o(x)$, and $p(x)=1/((\sigma_r^2)^{\text{agg}}(x))$. Define sums
\begin{equation}
\begin{aligned}
S_p   &= \sum_{x\in\mathcal{X}} p(x), \\
S_{pu} &= \sum_{x\in\mathcal{X}} p(x)u(x), \\
S_{pv} &= \sum_{x\in\mathcal{X}} p(x)v(x), \\
S_{puu} &= \sum_{x\in\mathcal{X}} p(x)u(x)^2, \\
S_{puv} &= \sum_{x\in\mathcal{X}} p(x)u(x)v(x).
\end{aligned}
\end{equation}
Then the minimizer is
\begin{equation}
a^\star=\frac{S_pS_{puv}-S_{pu}S_{pv}}{S_pS_{puu}-S_{pu}^2},\qquad
b^\star=\frac{S_{pv}-a^\star S_{pu}}{S_p}.
\label{eq:wls_closed}
\end{equation}

After calibration, we define residuals $\delta(x)=D_o(x)-\tilde{D}_r(x)$. Under the Gaussian model,
\begin{equation}
\delta(x)\sim \mathcal{N}\!\Bigl(0,\;\sigma_o^2+(a^\star)^2(\sigma_r^2)^{\text{agg}}(x)\Bigr).
\end{equation}
A simple maximum-likelihood (moment-matching) estimate of $\sigma_o^2$ is
\begin{equation}
\sigma_o^2=\max\!\left(0,\;\frac{1}{|\mathcal{X}|}\sum_{x\in\mathcal{X}}\left(\delta(x)^2-(a^\star)^2(\sigma_r^2)^{\text{agg}}(x)\right)\right).
\label{eq:sigo}
\end{equation}

Treating $D_o(x)$ as a Gaussian prior and $\tilde{D}_r(x)$ as a Gaussian likelihood, the product yields a Gaussian posterior with mean and variance:
\begin{align}
D_{\text{refined}}(x)
&=
\frac{
\frac{1}{\sigma_o^2}D_o(x)
+
\frac{1}{(a^\star)^2(\sigma_r^2)^{\text{agg}}(x)}\tilde{D}_r(x)
}{
\frac{1}{\sigma_o^2}
+
\frac{1}{(a^\star)^2(\sigma_r^2)^{\text{agg}}(x)}
},
\label{eq:bayes_fuse_mean} \\[6pt]
\sigma_{\text{refined}}^2(x)
&=
\left(
\frac{1}{\sigma_o^2}
+
\frac{1}{(a^\star)^2(\sigma_r^2)^{\text{agg}}(x)}
\right)^{-1}.
\label{eq:bayes_fuse_var}
\end{align}

If $\tau_r(x)=0$ (no NeRF support), we set $D_{\text{refined}}(x)=D_o(x)$ and $\sigma_{\text{refined}}^2(x)=\sigma_o^2$.

This fusion trusts the NeRF where $(\sigma_r^2)^{\text{agg}}$ is small, meaning ray termination distribution is sharp, which typically occurs near well-defined surfaces and discontinuities. MDENeRF reverts to the monocular prior in uncertain regions (e.g., disocclusions or diffuse density responses). Consequently, MDENeRF is aware of uncertainty while sharpening fine structure as opposed to heuristic edge weighting.

\subsection{Iterative Refinement Loop}
Each iteration outputs a refined depth $D_{\text{refined}}$ and uncertainty $\sigma_{\text{refined}}^2$, which initialize the next cycle. Bayesian fusion anchors refinement to the monocular prior and injects NeRF detail only where confidence is high, reverting elsewhere. Thus, our iterative refinement has minimal error accumulation; in practice, 2--3 iterations suffice.

\section{Experiments}
\textbf{Monocular Depth Baseline and Setup.} 
We use MiDaS (DPT-Large) for the initial monocular estimate $D_0$ in all our experiments~\cite{ranftl2022}. Since MiDaS produces scale-ambiguous depth, ground-truth depth is accessed only during evaluation via median scale alignment. The actual refinement steps of MDENeRF do not use ground-truth scale or metric supervision. We perform evaluations on 20 randomly selected indoor scenes on the SUN RGB-D dataset~\cite{song2015}.

\textbf{Implementation and Metrics.} 
We use small pose perturbations, generating $N{=}10$ synthetic views per scene. This preserves the original $640{\times}480$ resolution with minimal artifacts ($<2\%$ hole-filling). Within each scene, the NeRF is trained for 2000 optimization steps and refined for two iterations, as additional iterations result in diminishing returns. Although we use a low-fidelity NeRF, MDENeRF is plug-and-play and can improve from stronger NeRFs in future work. After median scale alignment, we report the global accuracy with mean squared error (MSE). High-frequency details are evaluated using edge sharpness (mean depth gradient magnitude, normalized relative to MiDaS) and edge F1. Additionally, we analyze our uncertainty quality with the correlation between predicted uncertainty and true depth error. For numerical stability, we clamp extremely small variances and skip affine calibration in degenerate cases, reverting to the monocular estimate.

\subsection{Quantitative Results}

Table~\ref{tab:main_quant} shows that MDENeRF consistently improves high-frequency depth structures while maintaining strong global accuracy. Compared to the MiDaS baseline, MDENeRF increases edge sharpness by $9\%$ and improves edge F1 by a relative amount of $2.9\%$. Together, these metrics indicate sharper and more accurate depth boundaries. Furthermore, global depth error is relatively maintained overall, with MDENeRF degrading by $1.92\%$ in MSE, indicating global error is largely preserved despite sharper boundaries. Ultimately, these results highlight the effectiveness of our fusion algorithm in refining MDE.

\begin{table}[h]
\centering
\caption{Average performance on SUN RGB-D test scenes. Global errors are reported after median scale alignment. Edge sharpness is normalized relative to MiDaS.}
\resizebox{\columnwidth}{!}{%
\begin{tabular}{lccc}
\hline
\textbf{Method} 
& \textbf{MSE} $\downarrow$ 
& \textbf{Edge Sharpness} $\uparrow$ 
& \textbf{Edge F1} $\uparrow$ \\
\hline
MiDaS (baseline) 
& $2.2835\times10^{3}$ 
& 1.00$\times$ 
& 0.414 \\

\textbf{MDENeRF (ours)} 
& $2.3274\times10^{3}$
& \textbf{1.09$\times$} 
& \textbf{0.426} \\
\hline
\end{tabular}}
\label{tab:main_quant}
\end{table}

\begin{figure}[h]
    \centering
    \includegraphics[width=1\linewidth]{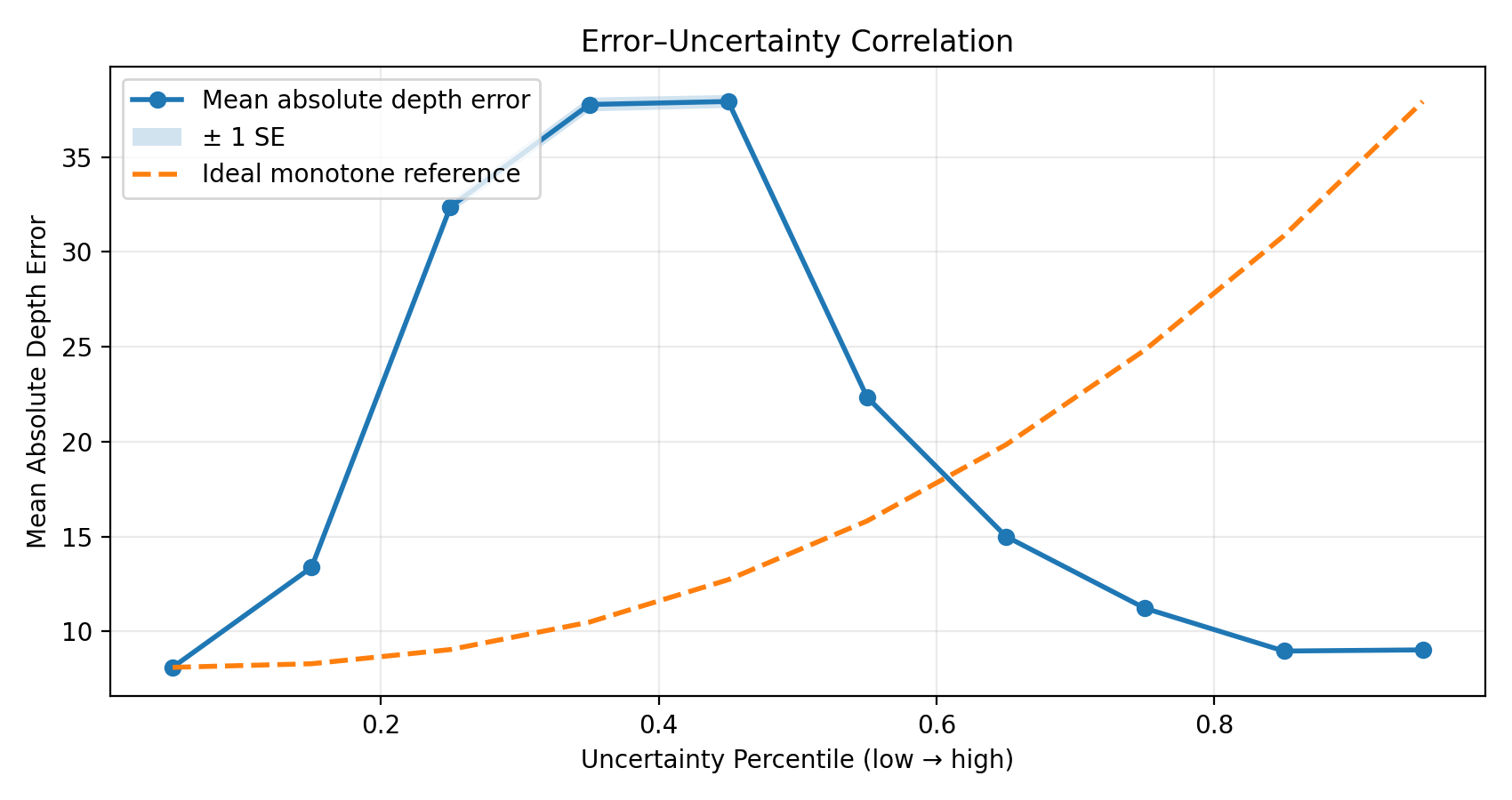}
    \caption{
    Correlation between predicted uncertainty and depth error.
    }
        \label{fig:error-uncertainty}
\end{figure}

Figure~\ref{fig:error-uncertainty} supports the formulations behind MDENeRF’s uncertainty estimates. After grouping pixels from low to high predicted uncertainty, it plots the mean absolute depth error (MAE) as a function of predicted uncertainty percentile. In the low-to-mid uncertainty regime, depth error increases with predicted uncertainty, indicating that regions MDENeRF labels as more uncertain tend to be less accurate. However, this trend does not hold across the entire uncertainty spectrum (Spearman $\rho=-0.038$), suggesting that predicted uncertainty is less informative in high-uncertainty regions. We attribute this behavior to the limited capacity of our lightweight NeRF model, which can trivially be improved in future work with stronger NeRFs.

\subsection{Qualitative Results}
\begin{figure}[h]
  \centering
  \includegraphics[width=\linewidth]{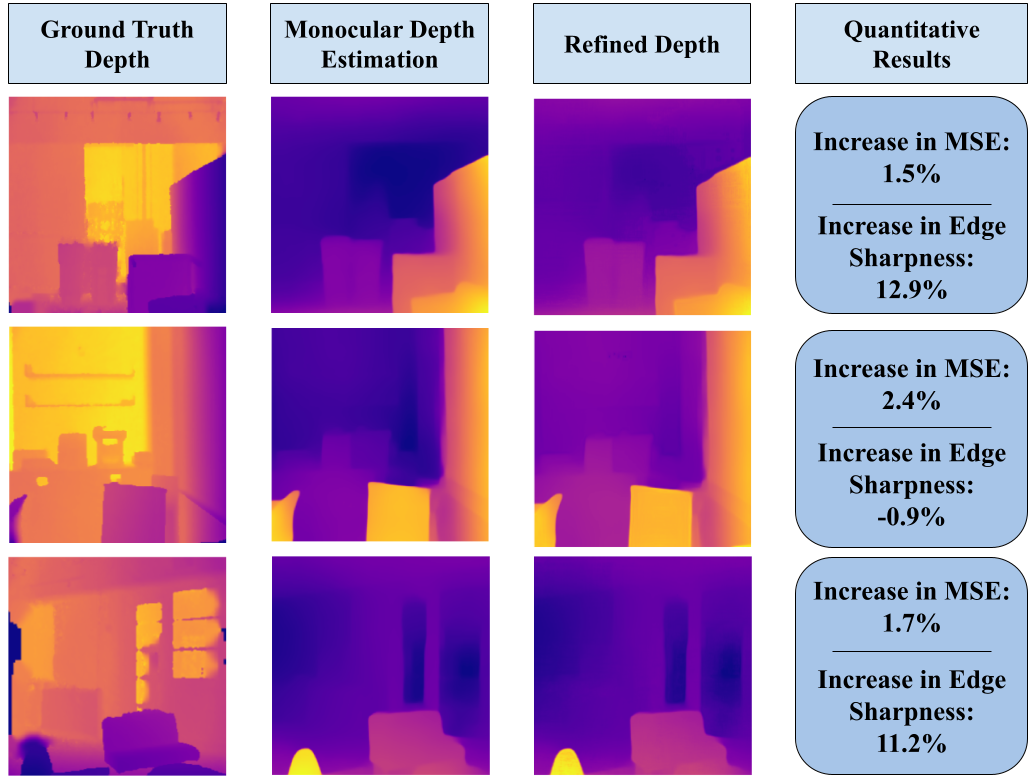}
  \caption{Qualitative results on challenging SUN RGB-D scenes. MDENeRF sharpens thin structures (e.g., chair legs and lamp poles) and strengthens occlusion boundaries while preserving planar regions such as walls and floors.}
  \label{fig:results}
\end{figure}

Figure~\ref{fig:results} contains challenging indoor scenes with thin objects, clutter, and sharp depth discontinuities. MiDaS estimates are globally consistent, but boundaries are visibly smooth, often thickening thin structures. On the other hand, MDENeRF achieves better separation in cluttered regions. The greatest improvements are near occlusion edges and fine structures due to sharper and less uncertain NeRF ray termination, which leads to a stronger influence during Bayesian fusion. In ambiguous areas, uncertainty increases, so the fusion reverts to our monocular prior.

\subsection{Ablation Studies}
Our fusion mechanism is closed-form and does not contain fusion parameters during inference. Therefore, we ablate our modeling choices rather than tuning parameters.

\begin{table}[h]
\centering
\caption{Ablation study of MDENeRF, where each variant removes a single component from the full model. Edge metrics are normalized relative to MiDaS. $\Delta$MSE denotes relative change compared to full MDENeRF.}
\begin{tabular}{lccc}
\hline
\textbf{Variant} 
& $\Delta$\textbf{MSE} $\downarrow$ 
& \textbf{Edge} $\uparrow$ 
& \textbf{F1} $\uparrow$ \\
\hline
Full MDENeRF 
& 0.0\% 
& 1.08$\times$ 
& 0.412 \\

-- NeRF variance 
& 0.5\% 
& 1.03$\times$ 
& 0.382 \\

-- Precision-weighted fusion 
& 0.7\% 
& 1.04$\times$ 
& 0.309 \\

-- Affine WLS calibration 
& 2.8\% 
& 1.06$\times$ 
& 0.402 \\

-- Empirical Bayes prior 
& 0.2\% 
& 1.08$\times$ 
& 0.397 \\

-- Monocular prior 
& 3.2\% 
& 1.10$\times$ 
& 0.406 \\
\hline
\end{tabular}
\label{tab:abl_fusion}
\end{table}

As shown in Table \ref{tab:abl_fusion}, we make three main observations after ablating.
(i) \emph{Uncertainty}: replacing NeRF variance with a constant reduces edge sharpness and degrades uncertainty-aware behavior, indicating that ray termination variance is a meaningful confidence signal.
(ii) \emph{Precision-weighted multi-view aggregation}: replacing our proposed fusion with heuristic minimum aggregation modestly degrades both global and local quality, especially in regions that are affected by disocclusion.
(iii) \emph{Calibration}: without the affine WLS alignment, global accuracy drops noticeably, indicating that the scale and shift must be corrected prior to probabilistic fusion. Finally, edge sharpness did slightly improve when removing the monocular prior, but global error significantly worsened, confirming its role in stabilizing refinement.

\section{Conclusions}

We present MDENeRF, a framework that iteratively refines monocular depth estimates via Bayesian fusion with NeRF depth. We interpret NeRF volume rendering weights as ray termination distributions, allowing for per-pixel depth uncertainty estimation. Fusing NeRF and monocular depth using Bayesian inference allows MDENeRF to selectively enhance high-frequency details while preserving the global scene structure. Our indoor experiments validate MDENeRF's fidelity in improving local accuracy, preserving global information, and calibrating uncertainty.

Current limitations of MDENeRF include scalability to larger scenes and more complex geometry. Furthermore, our NeRF training incurs computational costs that may be mitigated in future work. Potential directions involve multi-scale NeRFs and frequency-based analysis for more targeted refinement. Dynamic scene support is an open challenge, and such parallel lines of research may greatly support monocular depth refinement. Future work can readily integrate our plug-and-play framework into higher-fidelity NeRFs for improved uncertainty calibration and depth refinement. These advances in depth estimation will ultimately benefit safety-critical computer vision applications.

\bibliographystyle{IEEEtran}
\bibliography{main}

\end{document}